\documentclass{ieeeaccess}
\usepackage{cite}
\usepackage{amsmath,amssymb,amsfonts}
\usepackage{graphicx}
\usepackage{textcomp}

\usepackage{amsthm}
\usepackage{booktabs} 
\usepackage{multirow}
\usepackage{makecell}
\usepackage{algorithm,algpseudocode}
\usepackage{pifont}
\allowdisplaybreaks[1]

\newcommand{\be}{\begin{equation}}
\newcommand{\ee}{\end{equation}}
\newcommand{\ba}{\begin{eqnarray}}
\newcommand{\ea}{\end{eqnarray}}

\newcommand{\no}{\nonumber}

\newtheorem{theo}{Theorem}
\newcommand{\cmark}{\ding{51}}%
\newcommand{\xmark}{\ding{55}}%

\def\BibTeX{{\rm B\kern-.05em{\sc i\kern-.025em b}\kern-.08em
    T\kern-.1667em\lower.7ex\hbox{E}\kern-.125emX}}
\begin{document}
\history{This article has been accepted for publication in IEEE Access.}
\doi{10.1109/ACCESS.2022.3185095}

\title{RARTS: 
An Efficient First-Order Relaxed Architecture Search Method}
\author{\uppercase{Fanghui Xue}\authorrefmark{1}, 
\uppercase{Yingyong Qi\authorrefmark{1}, and Jack Xin}.\authorrefmark{1}}
\address[1]{Department of Mathematics, University of California at Irvine, Irvine, CA 92697, USA}
\tfootnote{
The work was partially supported by NSF grants DMS-1854434, DMS-1952644, and a Qualcomm Faculty Award.
}

\markboth
{F. Xue \headeretal: RARTS: 
An Efficient First-Order Relaxed Architecture Search Method}
{Author \headeretal: RARTS: 
An Efficient First-Order Relaxed Architecture Search Method}

\corresp{Corresponding author: Fanghui Xue (e-mail: fanghuix@uci.edu).}

\begin{abstract}
Differentiable architecture search (DARTS) is an effective 
method for data-driven neural network
design based on solving a bilevel optimization problem. Despite its success in many architecture search tasks, there are still some concerns about the accuracy of first-order DARTS and the efficiency of the second-order DARTS. In this paper, we formulate a single level alternative and a relaxed architecture search (RARTS) method that utilizes the whole dataset in architecture learning via both data and network splitting, without involving mixed second derivatives of the corresponding loss functions like DARTS. In our formulation of network splitting, two networks with different but related weights cooperate in search of a shared architecture. The advantage of RARTS over DARTS is justified by a convergence theorem and an analytically solvable model. Moreover, RARTS outperforms DARTS and its variants in accuracy and search efficiency, as shown in adequate experimental results. For the task of searching topological architecture, i.e., the edges and the operations, RARTS obtains a higher accuracy and 60\% reduction of computational cost than second-order DARTS on CIFAR-10. RARTS continues to out-perform DARTS upon transfer to ImageNet and is on par with recent variants of DARTS even though our innovation is purely on the training algorithm without modifying search space. For the task of searching width, i.e., the number of channels in convolutional layers, RARTS also outperforms the traditional network pruning benchmarks. Further experiments on the public architecture search benchmark like NATS-Bench also support the preeminence of RARTS.
\end{abstract}


\titlepgskip=-15pt

\maketitle

\section{Introduction}
Neural Architecture Search (NAS) is an automated machine learning technique to design an optimal neural network architecture by searching its building blocks of deep neural networks from a collection of candidate structures and operations. Although NAS has achieved many successes in several computer vision tasks \cite{zoph2016neural,zoph2018learning,ghiasi2019fpn,liu2019auto,elsken2019neural,ren2021comprehensive}, the search process demands  huge computational resources. The current search times have come down considerably 
from as many as 2000 GPU days in early NAS  \cite{zoph2018learning}, 
thanks to subsequent studies \cite{cai2018proxylessnas,liu2018progressive,pham2018efficient,real2019regularized,stamoulis2019single,wu2019fbnet, guo2020single} among others. Differentiable Architecture Search (DARTS) \cite{liu2019darts} is an appealing method that avoids searching over all possible combinations by relaxing the categorical architecture indicators to continuous parameters. The higher level architecture can be learned along with lower level weights via stochastic gradient descent by approximately solving a bilevel optimization problem. DARTS can be further sorted into first-order DARTS and second-order DARTS, in line with whether a mixed second derivative estimation of loss function is used or not. 

Despite its search efficiency obtained from continuous relaxation, DARTS can still have some problems experimentally and theoretically. They are the efficiency problem with second-order DARTS, the convergence problem with first-order DARTS, and the {\it architecture collapse} problem (i.e., the selected architecture contains too many skip-connections) with both DARTS.  Second-order DARTS takes much longer search time than first-order DARTS as it involves the mixed second derivatives. It has also been pointed out that second-order DARTS can have superposition effect \cite{he2020milenas}, which means the approximation of the gradient of $\alpha$ is based on the approximation of the weight $w$ one step ahead. This is believed to cause gradient errors and failures in finding optimal architectures. Therefore, it is used less often in practice than first-order DARTS \cite{he2020milenas,chen2019fasterseg}. However, first-order DARTS learns the architecture using half of the data only. Evidences are provided to show that it can result in incorrect limits and worse performance \cite{liu2019darts}. The experimental results also show that first-order DARTS (3.00\% error) is less accurate than second-order DARTS (2.76\% error) on the CIFAR-10 dataset \cite{krizhevsky2009learning, liu2019darts}.
For the architecture collapse problem, typically such a bias in operation selection degrades the model performance. This problem has been observed by a few researchers \cite{dong2019searching,chu2020fair}, who have tried to solve it by replacing some operations of the architecture. 

In addition to the search for topological architectures, i.e., edges and operations of cells (building blocks) in some early NAS works and DARTS \cite{zoph2018learning,liu2019darts}, many NAS style methods have been developed to search for the width of a model, i.e., the number of channels in convolutional layers \cite{dong2019network,chen2019fasterseg}. Searching for width is supposed to be a way of channel pruning, which is a common tool for {\it network compression}, i.e., constructing slim networks from redundant ones \cite{liu2017learning}. Specifically, channel pruning can be formulated as an architecture search problem, via the setup of learnable channel scoring parameters \cite{liu2017learning,huangwang2018,bui2021improving} as architecture parameters. This is an elegant approach for compression without relying on 
channel magnitude (group $\ell_1$ norm), which is used in previous regularization methods \cite{wen2016learning}. The previous way of setting up channel scoring parameters \cite{liu2017learning} utilizes the scale parameters of the batch normalization layers, yet they are not contained in many modern networks \cite{vaswani2017attention,dosovitskiy2020image}. Another challenge remains to be solved is to replace its plain gradient descent by the more accurate DARTS style algorithms.


Apart from the bilevel formulation of DARTS, a single level approach (SNAS) based on a differentiable loss and sampling has been proposed \cite{xie2018snas}. On CIFAR-10, SNAS is more accurate than the first-order DARTS yet with 50\% more search time than the second-order DARTS. This inspires us to formulate a new single level method which is more efficient and accurate.
Our {\it main contribution} is to introduce a novel {\it Relaxed Architecture Search (RARTS)} method based on single level optimization, and the computation of only the first-order partial derivatives of loss functions, for both topology and width search of architectures. Through both data and network splitting, the training objective (a relaxed Lagrangian function) of RARTS allows two networks with different but related weights to cooperate in the search of a shared architecture.

We have carried out both analytical and experimental studies below to show that RARTS achieves {\it better performance than first and second-order DARTS, with higher search efficiency than second-order DARTS} consistently:
\begin{itemize}

\item Compare RARTS with DARTS directly on the analytical model with quadratic loss functions, where the
RARTS iterations approach the true global minimal point missed by the first-order DARTS, in a robust fashion. A convergence theorem is proved for RARTS based on descent of its Lagrangian function, and equilibrium equations are discovered for the limits. 

\item On the CIFAR-10 based search of topological architecture, the model found by RARTS obtains smaller size and  higher test accuracy than that by the second-order DARTS with 65\% search time saving. 
A hardware-aware search option via a latency penalty in the Lagrangian function helps control the model size. Upon  transfer to ImageNet  \cite{deng2009imagenet,russakovsky2015imagenet}, the model found by RARTS  achieves better performance as well, compared with DARTS and its variants. Apart from the standard search space used in the DARTS paper, RARTS also beats DARTS on the public NAS benchmark of search spaces like NATS-Bench \cite{dong2021nats}.

\item  For channel pruning of ResNet-164 \cite{he2016deep} on CIFAR-10 and CIFAR-100 \cite{krizhevsky2009learning} with fixed pruning ratio (percentage of pruned channels), RARTS outperforms the differentiable pruning benchmarks: Network Slimming \cite{liu2017learning} and TAS \cite{dong2019network}. Comparisons between DARTS and RARTS have also been made in a $\ell_1$ regularized (unfixed ratio) pruning task, where RARTS achieves a high sparsity of 70\% and exceeds DARTS in accuracy. 

\end{itemize}

\section{Related work}

\subsection{Differentiable Architecture Search}\label{relate}
DARTS training relies on an iterative algorithm to solve a bilevel optimization problem \cite{Bilevel07,liu2019darts} which involves two loss functions computed via {\it data splitting} (splitting the dataset into two halves, i.e. training data and validation data):
\begin{align}\label{bilev}
\begin{split}
  & \mathop{\min}_{\alpha} L_{val}(w^{\ast}(\alpha), \alpha),\\
    \mathrm{where} \ \ \ \  &  w^{\ast}(\alpha) = \mathop{\arg\min}_w L_{train}(w, \alpha).
\end{split}
\end{align}
Here $w$ denotes the network weights, $\alpha$ is the architecture parameter, $L_{train}$ and $L_{val}$ are the loss functions computed on the training data $D_{train}$ and the validation data  $D_{val}$. Since many common datasets like CIFAR do not include the validation data, $D_{train}$ and $D_{val}$ are usually two non-overlapping halves of the original training data. We denote $L_{train}$ and $L_{val}$ by $L_{t}$ and $L_{v}$ to avoid any confusions with the meaning of the subscripts. $D_t$ and $D_v$ are defined similarly. DARTS has adopted data splitting because it is believed that joint training of both $\alpha$ and $w$ via gradient descent on the whole dataset by minimizing the overall loss function:
\begin{align}\label{nosplit}
  L(w, \alpha) =  L_{t}(w, \alpha) + L_{v}(w, \alpha)
\end{align}
can lead to overfitting \cite{liu2019darts,he2020milenas}.  Therefore, DARTS searches for the architectures through a two-step differentiable algorithm which updates the network weights and the architecture parameters in an alternating way:
\begin{itemize}
\item update weight $w$ by descending along $\nabla_w L_{t}(w,\alpha)$

\item update architecture parameter $\alpha$ by descending along:
\begin{align*}
\nabla_{\alpha}\, L_{v}(w - \xi \, \nabla_w L_{t}(w,\alpha),\alpha)
\end{align*}
\end{itemize}
where $\xi=0$ ($\xi > 0$ ) gives the first or second-order 
approximation. The bilevel optimization problem also arises in hyperparameter optimization and meta-learning, where a second-order algorithm and a convergence theorem on minimizers have been proposed in previous work \cite{Bilevel18} (Theorem 3.2), under the assumption that the $\alpha$-minimization is solved exactly, and $w^t(\alpha)$ converges uniformly to $w(\alpha)$. However, the $\alpha$-minimization of DARTS is approximated by gradient methods only, and hence the convergence of DARTS algorithm remains unknown theoretically.

We are aware of the fact that the first-order DARTS updates the architecture parameters on $D_v$ by descending along $\nabla_{\alpha}\, L_{v}(w,\alpha)$, which means it merely uses half of the data to train $\alpha$ and might cause some convergence issues (see Fig. \ref{convergence}). MiLeNAS has developed a mixed-level solution, where the architecture parameters can be learned on $D = D_{t} \cup D_{v}$ via a first-order descending algorithm \cite{he2020milenas}:
\begin{align}\label{IteMiLe}
\begin{split}
    w^{t+1} &= w^{t} - \eta^t_w \nabla_w  \, L_{t}(w^t, \alpha^t) \\
     \alpha^{t+1} &= \alpha^{t} - \eta^t_{\alpha}\,  \big(\nabla_{\alpha}L_{t}(w^{t+1}, \alpha^t) + \lambda \nabla_{\alpha}L_{v}(w^{t+1}, \alpha^t) \big),
     \end{split}
\end{align}
We shall see that MiLeNAS is actually a constrained case of RARTS when our two network splits become identical. However, we point out that computing $L_v$ using an identical network makes MiLeNAS still suffer from the same convergence issue in a later example (Section \ref{solvable}).
The second-order DARTS is observed to approximate the optimum better than first-order DARTS in a solvable model and through experiments, yet it requires computing the mixed derivative $\nabla^{2}_{\alpha,w} L_{t}$, at a considerable overhead. Searching by DARTS can also lead to the {\it architecture collapse} issue, meaning the selected architecture contains too many skip-connections. Typically such a bias in operation selection degrades the model performance. SNAS \cite{xie2018snas}, FBNet \cite{wu2019fbnet}, and GDAS \cite{dong2019searching} use differentiable Gumbel-Softmax to mimic one-hot encoding which implies exclusive competition and risks of unfair advantages \cite{chu2020fair}. This unfair dominance of skip-connections in DARTS has also been noted by FairDARTS \cite{chu2020fair}, which has proposed a collaborative competition approach by making the architecture parameters independent, through replacing the softmax with sigmoid. They have further penalized the operations in the search space with probability close to $\frac{1}{2}$, i.e. a neutral and ambiguous selection. As these methods focus on replacing some operations or the loss function, it would be worthwhile to explore other solutions such as replacing the gradient-based DARTS  search algorithm.

In addition to DARTS, many other differentiable methods for architecture search have been proposed, considering various aspects such as the search space, selection criterion, and training tricks. SNAS \cite{xie2018snas} has discussed it from a statistical perspective with however a minor to moderate performance improvement. The search efficiency has also been improved by sampling a portion of the search space during each update in training. A perturbation-based selection scheme has been proposed in \cite{wang2021rethinking}, as the magnitude of architecture parameters is believed to be inadequate as a selection criterion. P-DARTS \cite{chen2019progressive} has adopted operation dropout and regularization on skip-connections. From the procedure side to delay a quick short cut aggregation, it has also divided the search stage into multiple stages and progressively adds more depth than DARTS. PC-DARTS \cite{xu2019pc} samples a proportion of channels to reduce the bias of operation selection and enlarge the batch size as well.  GDAS \cite{dong2019searching} searches the architecture with one operation sampled at a time. 
Other approaches apply differentiable methods on much larger search spaces with sampling techniques to save memory and avoid model transfer \cite{cai2018proxylessnas, wu2019fbnet}. We will see that these variants of the differentiable architecture search method are actually complementary to our approach that advances DARTS on the purely algorithmic side by mobilizing weights. Moreover, many works  \cite{cai2018proxylessnas,he2018amc, wu2019fbnet, tan2019mnasnet,chen2019fasterseg,wan2020fbnetv2} manage to balance latency with the performance of the model to enhance the efficiency of the model. Despite the broad use of differentiable methods in the works we have mentioned, one may wonder how DARTS and its variants beat random search. A detailed comparison in \cite{li2020random} has elaborated the advantage of DARTS in accuracy and efficiency compared with random search.


\subsection{Search for width and channel pruning}\label{relateb}

Differentiable search method has contributed to a wide range of tasks other than topological architecture search.  TAS \cite{dong2019network} searches for the width of each layer, i. e. number of channels, by learning the optimal one from the aggregation of several candidate feature maps via a differentiable method and sampling. FasterSeg \cite{chen2019fasterseg} searches for the cell operations and layer width, as well as the multi-resolution network path over the semantic segmentation task. These works of searching width are closely related to channel pruning, which means pruning redundant channels from the convolutional layers. Among numerous methods to prune redundant channels \cite{hu2016network,wen2016learning,li2016pruning,he2017channel,luo2017thinet, liu2018rethinking}, a classical approach is to apply group LASSO \cite{yuan2006model} on the weights to identify unimportant channels.
The weights in each channel form one group, and the magnitude of each group is measured by  $\ell_2$ norm of its weights. The network is trained by minimizing a loss function penalized by the $\ell_1$ norm of these magnitudes  from all groups. 
The channels are pruned based on thresholding their norms. 
Selecting good thresholds as hyperparameters  
for different channels can be laborious for deep networks. 
On the other hand, 
channel selection is  intrinsically a network architecture issue. It is debatable if thresholding by weight magnitudes is always meaningful \cite{ye2018}.

Another approach of channel pruning \cite{liu2017learning,huangwang2018} involves assigning a channel scaling (scoring) factor to each channel, which is a learnable parameter independent of the weights. In the training process, the factors and the weights are learned jointly, and the channels with low scaling factors are pruned. After that, the optimal weights of the pruned network are adjusted by one more stage of fine-tuning. In terms of channel scaling factors, the channel pruning problem becomes a special case of neural architecture search.
Besides this formulation, there are several pruning methods based on NAS. AMC \cite{he2018amc} has defined a reward function and pruned the channels via reinforcement learning. MetaPruning \cite{liu2019metapruning} generates the best pruned model and weights from a meta network.


\section{Methodology}

In this section, we introduce the RARTS formulation, its iterative algorithm and convergence properties. RARTS is different from all the differentiable algorithms we have mentioned, in that it puts forward a relaxed formulation of a single level problem which benefits from both data splitting and network splitting.

\subsection{Data splitting and network splitting}\label{split}

As pointed out in DARTS \cite{liu2019darts} and MiLeNAS \cite{he2020milenas}, when learning the architecture parameter $\alpha$, splitting training and validation data should be taken into account to avoid overfitting. However, we have discussed that the bilevel formulation (\ref{bilev}) and training algorithm of DARTS may lead to several issues: unknown convergence, low efficiency and the unfair selection of operations. Therefore, we follow the routine of train-validation data splitting, but want to formulate a single level problem, in contrast to DARTS and MiLeNAS. First, if we use $(w, \alpha)$, the pair of weight and architecture parameters in Eq. (\ref{nosplit}) to represent a network, what we propose to do is to further relax the network weights $w$ via splitting a network copy denoted by $(y,\alpha)$. We call $(y, \alpha)$ and $(w,\alpha)$ the primary and auxiliary networks, which share the same architecture $\alpha$ and the same dimensions as weight tensors, but can have different weight initialization. 

Next, a primary loss $L_{v}(y, \alpha)$ is computed with parameters $(y,\alpha)$  fed on data $D_{v}$, while an auxiliary loss $L_{t}(w, \alpha)$ is computed with parameters $(w,\alpha)$ fed on data $D_{t}$. Note that the computation of the auxiliary loss $L_{t}(w, \alpha)$ is the same as that of DARTS. The difference is that the primary loss is computed on the primary network $(y, \alpha)$, instead of $(w,\alpha)$. Now we present the single level objective of our relaxed architecture search (RARTS) framework. With a $\ell_2$ penalty on the distance between $w$ and $y$, the two loss functions are combined through the following relaxed Lagrangian $L=L(y,w,\alpha)$ of Eq. (\ref{nosplit}):
\begin{align}\label{Lag}
    L := L_{v}(y, \alpha) + \lambda \, L_{t}(w, \alpha) + \frac{1}{2}\beta \, \|y-w\|^2_2,
\end{align}
where $\lambda$ and $\beta$ are hyperparameters controlling the penalty scale and the learning process. We will see in the search algorithm that the penalty term enables the two networks to exchange information and cooperate to search the architecture which they share together. This technique of splitting $w$ and $y$ is called {\it network splitting}, which is also inspired by some previous work \cite{dinh2020convergence}. In their work, splitting of variables is able to approximate a non-smooth minimization problem via an algorithm of combined closed-form solutions and gradient descent. 

Since various NAS approaches discover architectures of inconsistent sizes or FLOPS, it has made the comparison through different methods unfair, because larger models are likely to have better performance but low efficiency. Many NAS methods have adopted latency as a model constraint \cite{cai2018proxylessnas,chen2019fasterseg}.  To control model size, we follow the technique of approximating the model latency with the sum of latency from all the operations \cite{chen2019fasterseg}, and add the approximated latency to the loss function as a penalty. Since each component of the latency tensor (denoted by Lat) is the latency amount associated with a candidate operation, the dimension of Lat is the same as that of $\alpha$. Therefore, we provide an alternative objective which is penalized by the latency of the model:
\begin{align}\label{LatLag}
    L := & L_{v}(y, \alpha) + \lambda \, L_{t}(w, \alpha) + \frac{1}{2}\beta \, \|y-w\|^2_2 \nonumber \\
&   + \langle \mathrm{Softmax}(\alpha), \mathrm{Lat} \rangle, 
\end{align}
where the bracket is inner product. 

\subsection{RARTS Algorithm}\label{32}

We minimize the relaxed Lagrangian $L(y, w, \alpha)$ in (\ref{Lag}) by iteration on the three variables in an alternating way to allow individual and flexible learning schedules for the three variables. Similar to Gauss-Seidel method in numerical linear algebra \cite{Gauss}, we use updated variables immediately in each step and obtain the following three-step iteration:
\begin{align}\label{Ite}
\begin{split}
    w^{t+1} &= w^{t} - \eta^t_w\, \nabla_w L(y^t, w^t, \alpha^t) \\
    y^{t+1} &= y^{t} - \eta^t_y\, \nabla_y L(y^t, w^{t+1}, \alpha^t) \\ 
     \alpha^{t+1} &= \alpha^{t} - \eta^t_{\alpha}\, \nabla_{\alpha}L(y^{t+1}, w^{t+1}, \alpha^t).
\end{split}
\end{align}
With explicit gradient $\nabla_{w,y}\|y-w\|^2_2$, we have:
\begin{align}
\label{ite}
\begin{split}
    w^{t+1} =& \,w^{t}  -  \lambda\, \eta^t_w \nabla_w L_{t}(w^t, \alpha^t)-  \beta\eta^t_w(w^t-y^t) \\
    y^{t+1} =& \, y^{t}  - \eta^t_y\, \nabla_y L_{v}(y^t, \alpha^t)-  \beta\, \eta^t_y(y^t-w^{t+1}) \\ 
     \alpha^{t+1} =& \,\alpha^{t}  - \lambda \,\eta^t_{\alpha}\, \nabla_{\alpha}  L_{t}(w^{t+1}, \alpha^t)  \\
     &-   \eta^t_{\alpha}\, \nabla_{\alpha} L_{v}(y^{t+1}, \alpha^t).
\end{split}
\end{align}
To minimize the Lagrangian (\ref{LatLag}), the first two steps are the same as Eq. (\ref{ite}) since the latency only depends on $\alpha$. The third step becomes:
\begin{align}
\label{iteLat}
\begin{split}
     \alpha^{t+1} = & \alpha^{t}  - \lambda \, \eta^t_{\alpha}\, \nabla_{\alpha}  L_{t}(w^{t+1}, \alpha^t) \\
    &  -  \eta^t_{\alpha}\, \nabla_{\alpha} L_{v}(y^{t+1}, \alpha^t) \\
    & -\eta^t_{\alpha}\, \nabla_{\alpha} \langle \mathrm{Softmax}(\alpha^t), \mathrm{Lat} \rangle. 
\end{split}
\end{align}

Note that the update of $\alpha$ in Eq. (\ref{ite}) involves both 
$L_t$ and $L_v$, which is similar to the second-order DARTS but {\it without the mixed second derivatives.} The first-order DARTS uses $\nabla_{\alpha} L_{v}$ only in this step. In the previous section, we have discussed the architecture collapse issue of DARTS, i.e., selecting to many skip-connections. A possible reason why DARTS may lead to architecture collapse is that its architecture parameters converge more quickly than the weights in the convolutional layers. That means, when DARTS selects architecture parameters, it tends to select skip-connection operations, since the convolutional layers are not trained well. The fact that first-order DARTS has only used one of the two data splits  to train the weights, makes the training of convolutional layers worse. For RARTS, we make use of both $L_t$ and $L_v$ to update the weight parameters $w$ and $y$ in the first two steps of Eq. (\ref{ite}). In the third step of Eq. (\ref{ite}), both $L_t$ and $L_v$ are also used to update the shared architecture $\alpha$. In this way, the architecture is learned better, as more data are involved during training. If $y=w$ is enforced in Eq. (\ref{ite}) e.g. through a multiplier, RARTS essentially reduces to first-order MiLeNAS \cite{he2020milenas}. However, relaxing to $y \not = w$ has its advantages of having more generality and robustness as it is optimized on two networks with different but related weights. In contrast, MiLeNAS trains the network weigts on the training data $D_t$ only, and suffers from the same convergence issue as first-order DARTS (Section \ref{solvable}). We summarize the RARTS algorithm in Algorithm \ref{rarts0}.

\begin{algorithm}
\caption{Relaxed Architecture Search (RARTS)}\label{rarts0}
\hspace*{\algorithmicindent} \textbf{Input}: the number of iterations $N$, the hyperparameters $\lambda$ and $\beta$, a learning rate schedule $(\eta^t_w, \eta^t_u, \eta^t_{\alpha})$,     initialization of the weight parameters $w^0$, $u^0$ and the architecture parameters $\alpha^0$.\\ \hspace*{\algorithmicindent} \textbf{Output}: $\alpha^{\ast}$, the architecture we want.
\begin{algorithmic}
\State Split the dataset $D$ into two subsets $D_{p}$ and $D_{a}$.
\For{$t = 0, 1, ..., N$}
    \State Compute $L_{p}$ and $L_{a}$ on  $D_{p}$ and $D_{a}$, respectively, and then compute $L$ using Eq. (\ref{Lag})
    \State Update the parameters via gradient descent:
    \State $w^{t+1} = w^{t} - \eta^t_w\, \nabla_w L(y^t, w^t, \alpha^t)$
    \State $y^{t+1} = y^{t} - \eta^t_y\, \nabla_y L(y^t, w^{t+1}, \alpha^t)$
    \State $\alpha^{t+1} = \alpha^{t} - \eta^t_{\alpha}\, \nabla_{\alpha}L(u^{t+1}, w^{t+1}, \alpha^t)$
\EndFor
\end{algorithmic}
\end{algorithm}

  \begin{figure*}
      \centering
      {\includegraphics[width=1\textwidth]{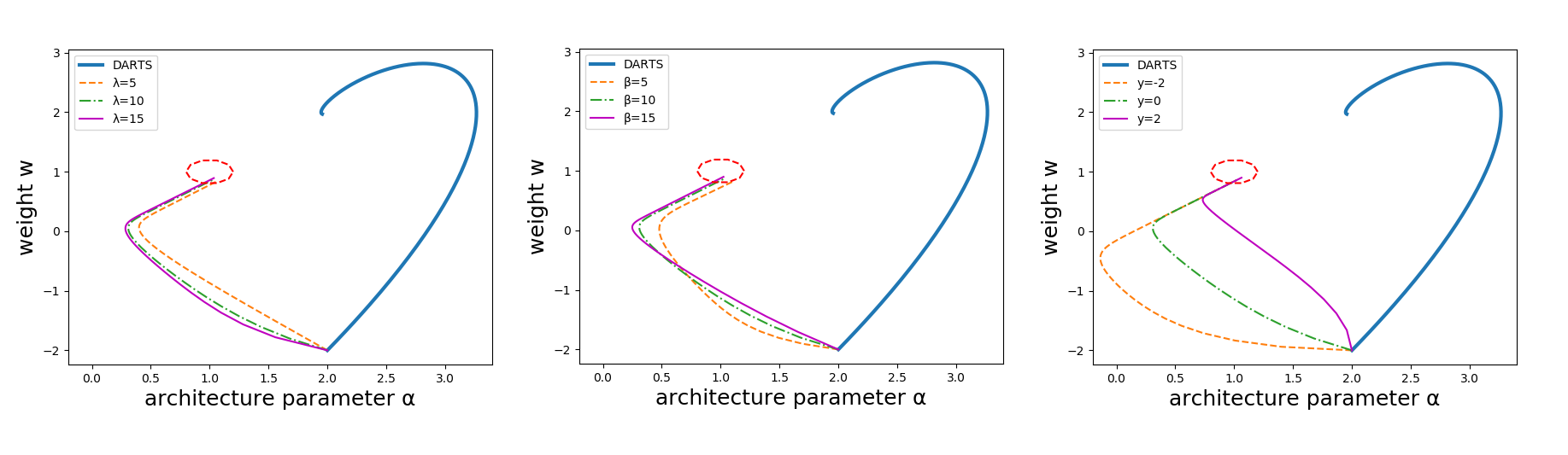}}%
    \caption{Learning trajectories of RARTS approach the global minimal point $(1,1)$ of the solvable model at suitable values of $\lambda$,  $\beta$  and $y_0$ ($\lambda=10$ in middle/right subplots, $\beta=10$ in left/right subplots, $y_0=0$ in left/middle subplots), compared with that of the baseline (first-order DARTS). 
    }
      \label{convergence}
    \end{figure*}

\subsection{Convergence analysis}
Suppose that $L_t$ and $L_v$ both satisfy Lipschitz gradient property, 
or there exist positive constants $L_1$ and $L_2$ such that ($z=(y,\alpha)$, $z'=(y',\alpha')$):
\[ \| \nabla_z L_v (z) - \nabla_z L_v (z') \| \leq L_{1}\|z - z'\|, \; \; \forall (z,z'), \]
which implies:
\[
L_v(z) -L_v(z') \leq  
\langle \nabla_{z} L_{v}(z'), (z-z') \rangle
+ \frac{L_{1}}{2}\|z-z'\|^2,
\]
for any $(z,z')$; 
similarly ($\zeta=(w,\alpha)$, $\zeta'=(w',\alpha')$): 
\[ \| \nabla_\zeta L_t (\zeta) - \nabla_\zeta L_t (\zeta') \| \leq L_{2}\|\zeta - \zeta'\|, \; \; \forall (\zeta,\zeta'), \]
which implies:
\[
L_t(\zeta) -L_{t}(\zeta') \leq 
  \langle\nabla_{\zeta}L_{t}(\zeta'), (\zeta-\zeta') \rangle
+\frac{L_2}{2}\|\zeta-\zeta'\|^2, 
\]
for any $(\zeta,\zeta')$. 

\begin{theo}\label{theo}
Suppose that the loss functions $L_t$ and $L_v$ satisfy  Lipschitz gradient property. If the learning rates $\eta_{w}^{t}$ 
, $\eta_{y}^{t}$ and $\eta_{\alpha}^{t}$ are small enough depending only on the Lipschitz constants as well as $(\lambda,\beta)$,  and approach nonzero limit at large $t$, the Lagrangian function  $L(y,w,\alpha)$ is descending on the iterations of (\ref{ite}). If additionally the Lagrangian $L$ 
is lower bounded and coercive (its boundedness implies that of its variables), the sequence $(y^t,w^t,\alpha^t)$ converges sub-sequentially to a critical point  $(\bar{y},\bar{w},\bar{\alpha})$ of $L(y,w,\alpha)$ obeying the equilibrium equations:
\ba
&&\lambda \nabla_{w} L_t (\bar{w},\bar{\alpha}) + \beta (\bar{w} -\bar{y}) = 0, \no \\
&& \nabla_y L_v(\bar{y},\bar{\alpha}) +\beta (\bar{y}-\bar{w})=0, \no \\
&& \lambda \nabla_\alpha L_t (\bar{w},\bar{\alpha}) + \nabla_{\alpha} L_v(\bar{y},\bar{\alpha}) =0. \label{equil}
\ea
If the loss is penalized by latency as in (\ref{LatLag}), the last equilibrium equation becomes:
\ba
\lambda \nabla_\alpha L_t (\bar{w},\bar{\alpha}) &+& \nabla_{\alpha} L_v(\bar{y},\bar{\alpha}) \nonumber \\ 
& +& \nabla_{\alpha} \langle 
  \mathrm{Softmax}(\bar{\alpha}),\mathrm{ Lat} \rangle=0. \label{equil_lat}
\ea
\end{theo}
\begin{proof}
We only need to prove for the loss (\ref{LatLag}) and the iterations (\ref{iteLat}), as the loss (\ref{Lag}) is its special case when $\mathrm{Lat}=0$. We notice the latency penalty function $\langle \mathrm{Softmax}(\alpha^t), \mathrm{Lat} \rangle$ also satisfies the Lipschitz gradient property. This is because 
\ba
    \nabla_{\alpha} \mathrm{Softmax}(\alpha^t) &=& \mathrm{diag}(\mathrm{Softmax}(\alpha^t)) \no \\
    & & - \mathrm{Softmax}(\alpha^t) \otimes (\mathrm{Softmax}(\alpha^t))',\no
\ea
and hence all the first and second derivatives of $\langle \mathrm{Softmax}(\alpha^t), \mathrm{Lat} \rangle$ are bounded uniformly regardless of $\alpha^t$.
Applying Lipschitz gradient inequalities on $L_v$ and $L_t$, we have:
 \ba
& & L(y^{t+1},w^{t+1},\alpha^{t+1}) - 
L(y^{t},w^{t},\alpha^{t}) \no \\
& = & L_v(y^{t+1},\alpha^{t+1}) + \lambda L_{t}(w^{t+1},\alpha^{t+1}) \no\\ 
& & +\frac{\beta}{2}\|y^{t+1}-w^{t+1}\|^2 + \langle \mathrm{Softmax}(\alpha^{t+1}), \mathrm{Lat} \rangle \no \\
& & -L_v(y^{t},\alpha^{t}) - \lambda L_{t}(w^{t},\alpha^{t}) 
 -\frac{\beta}{2}\|y^{t}-w^{t}\|^2 \no\\
 &&- \langle \mathrm{Softmax}(\alpha^t), \mathrm{Lat} \rangle \no \\
&\leq & \langle \nabla_{y,\alpha}\, L_v(y^t,\alpha^t), (y^{t+1}-y^{t},\alpha^{t+1}-\alpha^{t}) \rangle \no \\
& & +\frac{L_1}{2}\|(y^{t+1}-y^t,\alpha^{t+1}-\alpha^t)\|^2 \no\\
&&  + \lambda \, \langle \nabla_{w,\alpha}\, L_t(w^t,\alpha^t), (w^{t+1}-w^{t},\alpha^{t+1}-\alpha^{t}) 
\rangle \no \\
& & +\frac{L_2}{2}\|(w^{t+1}-w^t,\alpha^{t+1}-\alpha^t)\|^2  \no\\
&&+ \frac{\beta}{2}(\|y^{t+1}-w^{t+1}\|^2 
-\|y^{t}-w^{t}\|^2) \no \\
&& + \langle \nabla_{\alpha} \langle \mathrm{Softmax}(\alpha^t), \mathrm{Lat} \rangle, \alpha^{t+1}-\alpha^t) \rangle  \no\\
&& + \frac{L_3}{2}\|\alpha^{t+1}-\alpha^t\|^2. \no 
\ea
Substituting for the $(w,y)$-gradients from the  iterations (\ref{iteLat}), we continue:
\ba & & 
 L(y^{t+1},w^{t+1},\alpha^{t+1}) - 
L(y^{t},w^{t},\alpha^{t}) 
\no \\
& \leq & -(\eta_{y}^{t})^{-1}\, \no\\
&& \cdot\langle y^{t+1}-y^t +\beta \, \eta_{y}^{t}\, (y^t - w^{t+1}), y^{t+1}-y^t \rangle \no \\
& & + \langle \nabla_\alpha L_v (y^t,\alpha^t) + \lambda \nabla_\alpha L_t (w^t,\alpha^t), \alpha^{t+1}-\alpha^t \rangle  \no \\
&& +\lambda\,(-\lambda \eta_{w}^{t})^{-1} \no\\
&&\cdot \langle w^{t+1} - w^t    +\beta \, \eta_{w}^{t} \, (w^t - y^t), w^{t+1}-w^t \rangle  \no \\
&&  +\frac{L_1}{2}\|y^{t+1}-y^t\|^2 
+\frac{L_1+L_2+L_3}{2}\|\alpha^{t+1}-\alpha^t\|^2  \no\\
&&+ \frac{L_2}{2}\|w^{t+1}-w^t\|^2  \no \\
&& + \frac{\beta}{2}(\|y^{t+1}-w^{t+1}\|^2-\|y^{t}-w^{t}\|^2)  \no \\
&&+ \langle \nabla_{\alpha} \langle \mathrm{Softmax}(\alpha^t), \mathrm{Lat} \rangle, \alpha^{t+1}-\alpha^t) \rangle  \no \\
&= & (-(\eta_{y}^{t})^{-1}+L_{1}/2) \, \|y^{t+1}-y^t\|^2 \no\\ && +(-(\eta_{w}^{t})^{-1}+L_2/2)\, \|w^{t+1}-w^t \|^2  \no \\
& & -\beta \langle y^t - w^{t+1}, y^{t+1}-y^t \rangle \no\\
&&- \beta \langle w^t-y^t, w^{t+1} -w^t \rangle  \no \\
& &+ \frac{\beta}{2}(\|y^{t+1}-w^{t+1}\|^2 
-\|y^{t}-w^{t}\|^2) \no\\
&&+ \langle \nabla_\alpha L_v (y^t,\alpha^t)
+ \lambda \nabla_\alpha L_t (w^t,\alpha^t), \alpha^{t+1}-\alpha^t \rangle \no \\
& & + \frac{L_1+L_2+L_3}{2}\|\alpha^{t+1}-\alpha^t\|^2  \no\\
&&+ \langle \nabla_{\alpha} \langle \mathrm{Softmax}(\alpha^t), \mathrm{Lat} \rangle, \alpha^{t+1}-\alpha^t) \rangle. \label{prf1}
\ea

We note the following identity
\ba
 && \|y^{t+1}-w^{t+1}\|^2 \no \\
& = & \|y^{t+1}-w^t + w^t -w^{t+1}\|^2 \no \\
& = & \|y^{t+1} - w^t \|^2 + 2 \langle y^{t+1}-w^t, w^t -w^{t+1} \rangle \no\\
&&+\|w^t - w^{t+1}\|^2, \no
\ea
where
\begin{align*}
    &\|y^{t+1} - w^t \|^2\\
    =\quad  &\|-w^t + y^t -y^t + y^{t+1} \|^2\\
    =\quad  &\|y^t - w^t\|^2 + 2 \langle y^t -w^t, y^{t+1}-y^t \rangle + \|y^{t+1}-y^t\|^2.
\end{align*}
Upon substitution of the above 
in the right hand side of (\ref{prf1}), we find that:
\ba & & 
L(y^{t+1},w^{t+1},\alpha^{t+1}) - 
L(y^{t},w^{t},\alpha^{t}) 
\no \\
& \leq &  (-(\eta_{y}^{t})^{-1}+L_{1}/2+\beta/2) \, \|y^{t+1}-y^t\|^2 \no\\ &&+(-(\eta_{w}^{t})^{-1}+L_2/2 +\beta/2)\, \|w^{t+1}-w^t \|^2  \no \\
&& + \beta \langle w^{t+1} -w^t, y^{t+1}-y^t \rangle \no\\
&&+ \beta \langle y^{t+1}-y^t, w^t -w^{t+1} \rangle \no  \\
& &+  
\langle \nabla_\alpha L_v (y^t,\alpha^t)
+ \lambda \nabla_\alpha L_t (w^t,\alpha^t), \alpha^{t+1}-\alpha^t \rangle \no\\
&&+ \frac{L_1+L_2+L_3}{2}\|\alpha^{t+1}-\alpha^t\|^2 \no \\
& & + \langle \nabla_{\alpha} \langle \mathrm{Softmax}(\alpha^t), \mathrm{Lat} \rangle, \alpha^{t+1}-\alpha^t) \rangle. \label{prf2}
\no
\ea
The $\beta$-terms cancel out. 
Substituting for the $\alpha$-gradient from the iterations (\ref{iteLat}), we get:
\ba
& & L(y^{t+1},w^{t+1},\alpha^{t+1}) - 
L(y^{t},w^{t},\alpha^{t}) \no \\
&\leq &  (-(\eta_{y}^{t})^{-1}+L_{1}/2+\beta/2) \, \|y^{t+1}-y^t\|^2 \no\\ && +(-(\eta_{w}^{t})^{-1}+L_2/2 +\beta/2)\, \|w^{t+1}-w^t \|^2  \no \\
& & + (-(\eta_{\alpha}^{t})^{-1} + \frac{L_1+L_2+L_3}{2})\|\alpha^{t+1} -\alpha^{t}\|^2 \no \\
& & +\langle \nabla_{\alpha} L_v(y^t,\alpha^t) -\nabla_{\alpha} L_v(y^{t+1},\alpha^{t}), \alpha^{t+1}-\alpha^t \rangle \no \\
& & +\lambda \langle 
\nabla_{\alpha} L_t(w^t,\alpha^t) 
-\nabla_{\alpha} L_t (w^{t+1},\alpha^t),\alpha^{t+1}-\alpha^t \rangle \no 
\ea
where the last two inner product terms are upper bounded by:
\[ (1+\lambda)\, L_4\,  (\|y^t - y^{t+1} \|+\|w^t -w^{t+1}\|)\, \|\alpha^{t+1} -\alpha^t\|, \]
for positive constant $L_4:= \max(L_1,L_2)$. 
It follows that:
\begin{align}
&L(y^{t+1},w^{t+1},\alpha^{t+1}) - 
L(y^{t},w^{t},\alpha^{t}) \no \\
\leq &\left[-(\eta_{y}^{t})^{-1}\!+\!\frac{L_{1}}{2}\!+\!\frac{\beta}{2}\!+\!(1\!+\!\lambda)\, \frac{L_4}{2}\right] \, \|y^{t+1}-y^t\|^2\no \\
&+\left[-(\eta_{w}^{t})^{-1}\!+\!\frac{L_2}{2}\!+\!\frac{\beta}{2}\!+\!(1\!+\!\lambda)\, \frac{L_4}{2}\right]\|w^{t+1}-w^t \|^2  \no \\
&+\left[-(\eta_{\alpha}^{t})^{-1}\!+\frac{L_1+L_2+L_3}{2} + \!(1\!+\!\lambda)\, \frac{L_4}{2}\right]\, \no\\
&\cdot \|\alpha^{t+1} -\alpha^{t}\|^2. \label{descent}
\end{align}

If 
\begin{align*}
    \eta^{t}_{y} &< \frac{1}{2} \left[\frac{L_{1}}{2}+\frac{\beta}{2} +(1+\lambda)\, \frac{L_4}{2}\right]^{-1}:=c_1,\\
    \eta_{w}^{t} &< \frac{1}{2} \left[\frac{L_2}{2} +\frac{\beta}{2}+ (1+\lambda)\, \frac{L_4}{2}\right]^{-1}:=c_2,\\
    \eta^{t}_{\alpha} &<  \frac{1}{2}\left[\frac{L_1+L_2+L_3}{2} + \!(1\!+\!\lambda)\, \frac{L_4}{2}\right]^{-1}:=c_3,
\end{align*} 
$L$ is descending along the sequence 
$(y^t,w^t,\alpha^t)$.
For $c_4 = \frac{1}{2}\min\{c_{1}^{-1},c_{2}^{-1},c_{3}^{-1}\}$, it follows from (\ref{descent})
that:
\begin{align*}
    & c_4\|(y^{t+1}-y^t,w^{t+1}-w^t,\alpha^{t+1}-\alpha^t )\|^2\\
\leq & \,L(y^{t},w^{t},\alpha^{t}) -
L(y^{t+1},w^{t+1},\alpha^{t+1}) \rightarrow 0
\end{align*}
as $ t \rightarrow +\infty$, implying that 
\[ \lim_{t \rightarrow \infty} \| (y^{t+1}-y^t,w^{t+1}-w^t,\alpha^{t+1}-\alpha^t )\| = 0.\]
Since $L$ is lower bounded and coercive, 
$\|(y^t,w^t,\alpha^t)\|$ are uniformly bounded in $t$. Let $(\eta_{w}^{t}, \eta_{y}^{t}, \eta_{\alpha}^{t})$ tend to non-zero limit at large $t$.
Then $(y^t,w^t,\alpha^t)$ sub-sequentially converges to a limit point 
$(\bar{y},\bar{w},\bar{\alpha})$ satisfying the equilibrium system  (\ref{equil}) or (\ref{equil_lat}).
\end{proof}

\subsection{A solvable bilevel model}\label{solvable}
We compare a few differentibale methods through an example \cite{liu2019darts} which has an analytical solution. Regardless of the latency penalty, we consider quadratic functions
$L_{v}=\alpha\,w -2\alpha +1$, $L_{t}=w^2 -2\, \alpha \,w +\alpha^2$ for the bilevel problem (\ref{bilev}). Therefore, the solution to the inner level problem is:
\begin{align*}
    w^{\ast}(\alpha) = \mathop{\arg\min}_w L_t(w, \alpha) = \alpha.
\end{align*}
Then $L_{v}(w^{\ast}(\alpha), \alpha) = 
\alpha^2 -2\alpha +1$, and the global minimizer of this bilevel problem is $(w^{\ast},\alpha^{\ast})=(1,1)$. However, the equilibrium equations of first-order DARTS is:
\begin{align*}
\begin{split}
    & \nabla_w L_{t}(\bar{w}, \bar{\alpha}) = 0  \\
    & \nabla_{\alpha} L_{v}(\bar{w}, \bar{\alpha}) = 0,
    \end{split}
\end{align*}
which gives a spurious equilibrium $(\bar{w}, \bar{\alpha}) = (2,2)$. The equilibrium equations of first-order MiLeNAS is:
\begin{align*}
\begin{split}
    & \nabla_w L_{t}(\bar{w}, \bar{\alpha}) = 0  \\
    & \nabla_{\alpha}L_{t}(\bar{w}, \bar{\alpha}) +  \lambda \nabla_{\alpha}L_{v}(\bar{w}, \bar{\alpha}) = 0,
    \end{split}
\end{align*}
which also results in the spurious equilibrium $(\bar{w}, \bar{\alpha}) = (2,2)$. 

On the other hand, RARTS can approximate the correct minimizer $(w^{\ast},\alpha^{\ast})=(1,1)$ better. Note that both $L_v$ and $L_t$ satisfy the Lipschitz gradient property, which implies the descent of Lagrangian $L$ by the proof of Theorem \ref{theo}. If $\lambda > 1/2$, $\beta > 3/2$, $L$ is bounded and coercive, which follows from an eigenvalue analysis of linear system (\ref{ite}) and is observed in computation. 
Hence, Theorem \ref{theo} can be applied to this example, and the equilibrium system  (\ref{equil}) reads:
\ba
\lambda (2\bar{w} - 2\bar{\alpha}) + \beta (\bar{w} - \bar{y}) &=& 0, \label{elib1}\\
\bar{\alpha} + \beta (\bar{y} -\bar{w}) &=& 0, \label{elib2}\\
 \lambda (-2 \bar{w} + 2 \bar{\alpha}) + \bar{y} - 2 &=& 0. \label{elib3}
\ea
Adding (\ref{elib1}) and (\ref{elib2}) gives: $\bar{w} = \frac{2\lambda -1}{2 \lambda } \bar{\alpha}$, which together with  (\ref{elib3}) determines $(\bar{\alpha}, \bar{w}, \bar{y})$ uniquely: $(\bar{\alpha},\bar{w}, \bar{y})= (\frac{4\beta\lambda}{4\beta\lambda - \beta -2\lambda},\frac{4\beta\lambda-2\beta}{4\beta\lambda - \beta -2\lambda},\frac{4\beta\lambda-2\beta-4\lambda}{4\beta\lambda - \beta -2\lambda})$, if $4\beta\lambda - \beta -2\lambda \neq 0$. At $\lambda=\beta=15$, $(\bar{\alpha},\bar{w},\bar{y})\approx (1.053,1.018,0.947)$ where {\it global convergence holds for the whole RARTS sequence.} The learning dynamics starting from $(\alpha_0,w_0,y_0)=(2,-2,y_0)$, is reproduced in Fig. \ref{convergence}, along with three learning curves from RARTS as the parameters $(\lambda, \beta)$ and the initial value $y_0$
vary. In Fig. \ref{convergence}a, $\beta = 10$, $y_0=0$. 
In Fig. \ref{convergence}b, $\lambda=10$, $y_0=0$. In Fig. \ref{convergence}c, $\lambda=\beta =10$. In all experiments, the learning rates are fixed at $0.01$. For a  range of $(\lambda,\beta)$ and $y_0$, we see that our learning curves enter a small circle around $(1,1)$, while first-order DARTS converges to the spurious point.

\begin{table*}[t]
        \begin{center}
            \caption{Comparison of DARTS,
            RARTS and other methods on CIFAR-10 based network search. DARTS-1/2 stands for DARTS 1st/2nd-order, SNAS-Mi/Mo stands for SNAS plus mild/moderate constraints.
    Note that faster search times also depend on speed and memory capacity of local machines used. 
    The V100 column indicates whether the model is trained on high-end Tesla V100 GPUs or not.
    Each run of our experiment is conducted on a single GTX 1080 Ti GPU.
    The numbers in the parentheses indicate the search GPU days of DARTS on our machine. 
    Average of 5 runs. $\diamond$ These runs are conducted on our machine. }    \label{NAS}
    \begin{tabular}{l c c c c c}
    \toprule
     \specialrule{0em}{1pt}{1pt}
    \multirow{2}{*}{\makecell{Method}} &
    \multirow{2}{*}{\makecell{Test Error (\%)}} &\multirow{2}{*}{\makecell{Para. (M)}} &\multirow{2}{*}{\makecell{V100}}
    & Search 
    \\ 
    & & & & GPU Days
    \\ 
    \specialrule{0em}{1pt}{1pt}
    \midrule
     \specialrule{0em}{1pt}{1pt}
          Random Baseline \cite{liu2019darts} & 3.29 $\pm$ 0.15 & 3.2 & \xmark & 4 \\
    \specialrule{0em}{1pt}{1pt}
    \midrule
     \specialrule{0em}{1pt}{1pt}
    AmoebaNet-B \cite{real2019regularized} & 2.55 $\pm$ 0.05 & 2.8 & \xmark & 3150\\
    SNAS-Mi \cite{xie2018snas} & 2.98 & 2.9 & \xmark & 1.5 \\
     SNAS-Mo \cite{xie2018snas} & 2.85 $\pm$ 0.02 & 2.8 & \xmark & 1.5 \\
    DARTS-1 \cite{liu2019darts} & 3.00 $\pm$ 0.14 & 3.3 & \xmark & 1.5 (0.7)  \\
     DARTS-2 \cite{liu2019darts} & 2.76 $\pm$ 0.09 & 3.3 & \xmark & 4 (3.1)  \\
            \specialrule{0em}{1pt}{1pt}
      \midrule
     \specialrule{0em}{1pt}{1pt}    GDAS \cite{dong2019searching} & 2.82  & 2.5 & \cmark & 0.2   \\
    ProxylessNAS \cite{cai2018proxylessnas} & 2.08  & 5.7 & \cmark &  4.0\\ 
    FairDARTS \cite{chu2020fair} & 2.54 $\pm$ 0.05  & 3.3 & \cmark
     & 0.4 \\
    FairDARTS \cite{chu2020fair} & 2.94 $\pm$ 0.05 $\diamond$ & 3.2
     & \xmark & 0.3 \\
    P-DARTS \cite{chen2019progressive} & 2.50  & 3.4  &  \cmark & 0.3\\
    PC-DARTS \cite{xu2019pc}  & 2.57 $\pm$ 0.07 & 3.6 & \cmark & 0.1 \\
    PC-DARTS \cite{xu2019pc}  & 2.71 $\diamond$ & 2.9  & \xmark & 0.1 \\
    MiLeNAS \cite{he2020milenas} & 2.80 $\pm$ 0.04   & 2.9 & \cmark & 0.3  \\
    MiLeNAS \cite{he2020milenas} & 2.51 $\pm$ 0.11  & 3.9 & \cmark & 0.3   \\
      \specialrule{0em}{1pt}{1pt}
      \midrule
     \specialrule{0em}{1pt}{1pt}
    RARTS 
    & 2.65 $\pm$ 0.07 & 3.2 &\xmark &  1.1  \\    
     \specialrule{0em}{1pt}{1pt}
    \bottomrule
    \end{tabular}
    \end{center}
\end{table*}

\begin{table*}[t]
        \begin{center}
            \caption{Comparison of the latency for the models found under different hyperparameters. The setting of batch size = 64, learning rate = $3\times10^{-4}$, weight decay = $1\times10^{-3}$ is consistent with the settings of DARTS and other DARTS variants, and is selected to be our baseline setting.}    \label{hyperlat}
    \begin{tabular}{c c c c c}
    \toprule
     \specialrule{0em}{1pt}{1pt}
    Latency Weight & Batch Size & Learning Rate & Weight Decay & Latency (ms)\\
      \specialrule{0em}{1pt}{1pt}
      \midrule
     \specialrule{0em}{1pt}{1pt}
    $2\times10^{-3}$ & 64 & $3\times10^{-4}$ & $1\times10^{-3}$ & 21.7\\
    $2\times10^{-2}$ & 64 & $3\times10^{-4}$ & $1\times10^{-3}$ & 12.4\\
    $2\times10^{-4}$ & 64 & $3\times10^{-4}$ & $1\times10^{-3}$ & 23.4\\
      \specialrule{0em}{1pt}{1pt}
      \midrule
     \specialrule{0em}{1pt}{1pt}
    $2\times10^{-3}$ & 64 & $3\times10^{-4}$ & $1\times10^{-3}$ & 21.7\\
    $2\times10^{-3}$ & 64 & $3\times10^{-3}$ & $1\times10^{-3}$ & 21.0\\
    $2\times10^{-3}$ & 64 & $3\times10^{-5}$ & $1\times10^{-3}$ & 23.1\\
      \specialrule{0em}{1pt}{1pt}
      \midrule
     \specialrule{0em}{1pt}{1pt}
    $2\times10^{-3}$ & 64 & $3\times10^{-4}$ & $1\times10^{-3}$ & 21.7\\
    $2\times10^{-3}$ & 64 & $3\times10^{-4}$ & $1\times10^{-4}$ & 21.3\\
    $2\times10^{-3}$ & 64 & $3\times10^{-4}$ & $1\times10^{-2}$ & 22.9\\
      \specialrule{0em}{1pt}{1pt}
      \midrule
     \specialrule{0em}{1pt}{1pt}
    $2\times10^{-3}$ & 64 & $3\times10^{-4}$ & $1\times10^{-3}$ & 21.7\\
    $2\times10^{-3}$ & 32 & $3\times10^{-4}$ & $1\times10^{-3}$ & 20.1\\  
    $2\times10^{-3}$ & 16 & $3\times10^{-4}$ & $1\times10^{-3}$ & 16.5\\
     \specialrule{0em}{1pt}{1pt}
    \bottomrule
    \end{tabular}
    \end{center}
\end{table*}

\section{Experiments}
We show by a series of experiments how RARTS works efficiently for
different tasks: the search for topology and the search for width, on various datasets and search spaces.
\subsection{Search for Topology}

For the hyperparameters and settings like learning rate schedules, number of epochs for CIFAR-10 and the transfer learning technique for ImageNet, we follow those of DARTS \cite{liu2019darts}. We also consider the results on CIFAR-10 and CIFAR-100 for NATS-Bench \cite{dong2021nats}, which is another benchmark search space. 


    \begin{table*}[t]
        \begin{center}
          \caption{Transfer to ImageNet: test error comparison of DARTS, RARTS and other methods on local machines resp. The V100 column indicates whether the model is trained on high-end Tesla V100 GPUs or not. The larger GPU memory can support larger batch size, which leads to better accuracy and training efficiency on ImageNet. 
        The Direct column indicates if the model is searched directly on ImageNet without transfer-learning. The direct search tends to be more accurate but costs more computational resources.   }  \label{Ima}
    \begin{tabular}{l l c c c c}
    \toprule
     \specialrule{0em}{1pt}{1pt}
    Method & Top-1 (\%) &
    Top-5 (\%) & Parameters (M) & V100 & Direct \\
    \specialrule{0em}{1pt}{1pt}
    \midrule
     \specialrule{0em}{1pt}{1pt}
     SNAS 
     \cite{xie2018snas} 
     & 27.3 & 9.2 & 4.3 & \xmark & \xmark \\
    DARTS \cite{liu2019darts}
      & 26.7 & 8.7 & 4.7 & \xmark & \xmark \\
      \specialrule{0em}{1pt}{1pt}
      \midrule
     \specialrule{0em}{1pt}{1pt}
     GDAS 
     \cite{dong2019searching}
     & 26.0  & 8.5 & 5.3 & \cmark & \xmark \\
  ProxylessNAS \cite{cai2018proxylessnas} & 24.9   & 7.5 & 7.1 & \cmark & \cmark\\ 
    FairDARTS  \cite{chu2020fair} &
    24.9  & 7.5 & 4.8 & \cmark & \xmark \\
    FairDARTS  \cite{chu2020fair}   & 
    24.4  & 7.4 & 4.3 & \cmark & \cmark \\
    P-DARTS \cite{chen2019progressive} &  24.4  & 7.4 & 4.9 & \cmark & \xmark \\
    PC-DARTS \cite{xu2019pc} & 25.1 & 7.8 & 5.3 & \cmark & \xmark \\
    PC-DARTS \cite{xu2019pc} & 24.2 & 7.3 & 5.3 & \cmark & \cmark \\
    MiLeNAS
     \cite{he2020milenas}
     & 25.4 & 7.9 & 4.9 & \cmark & \xmark \\
      \specialrule{0em}{1pt}{1pt}
      \midrule
     \specialrule{0em}{1pt}{1pt}
    RARTS 
    & 25.9 & 8.3 & 4.7 & \xmark & \xmark \\    
     \specialrule{0em}{1pt}{1pt}
    \bottomrule
    \end{tabular}
    \end{center}
\end{table*}

\begin{figure}
      \centering
      {\includegraphics[width=0.48\textwidth]{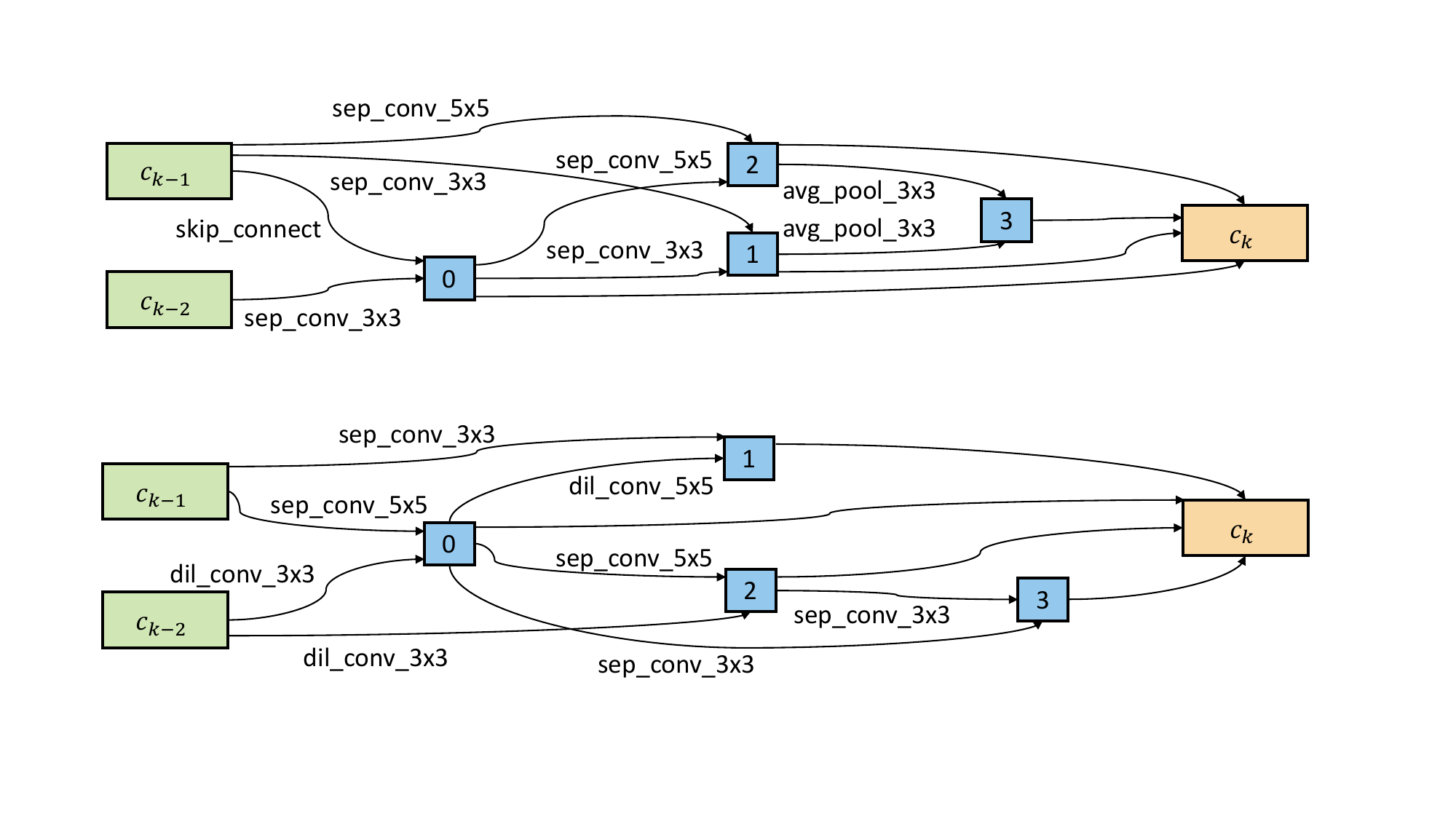}}
    \caption{The architecture of the normal (top) and reduction (bottom) cells found by RARTS. This architecture contains only one skip connection. The last four edges are simply concatenated together to construct the next cell. So there is no search along these edges, following the convention of DARTS  \cite{liu2019darts}.} 

      \label{arch}
    \end{figure}

\textbf{Comparisons on CIFAR-10.} 
The CIFAR-10 dataset consists of 50,000 training images and 10,000 test images \cite{krizhevsky2009learning}. These 3-channel images of $32\times32$ resolutions are allocated to 10 object classes evenly. For the architecture search task on CIFAR-10, the $D_t$ and $D_v$ data we have used are random non-overlapping halves of the original training data, the same as DARTS. The settings for searching topology with RARTS follows those of DARTS. That is, batch size = 64, initial weight learning rate = 0.025, momentum = 0.9, weight decay = 0.0003, initial alpha learning rate = 0.0003, alpha weight decay = 0.001, epochs = 50. For the stage of training, batch size = 96, learning rate = 0.025, momentum = 0.9, weight decay = 0.0003 \cite{liu2019darts}. For each cell (either normal or reduction), 8 edges are selected, with 1 out of 8 candidate operations selected for each edge (see Fig. \ref{arch}). Besides the standard $\ell_2$ regularization of the weights, we also adopt the latency penalty. The latency regularization loss is {\it weighted} so that it is balanced with other loss terms. Typically, if we increase the latency weight, the model we find will be smaller in size. The latency term Lat for each operation is measured via PyTorch/TensorRT \cite{chen2019fasterseg}, and thus it depends on the devices we use. 
For the current search, the latency weight is 0.002 so that the model size is comparable to those in prior works. The final latency loss is the weighted sum of the latency from each operation, where the weights are the architecture parameters. 

As shown in Table \ref{NAS}, the search cost of RARTS is $1.1$ GPU days, far less than that of the second-order DARTS. The test error of RARTS is $2.65\%$, outperforming the $3.00\%$ of the first-order DARTS and the $2.76\%$ of the second-order DARTS. It should also be pointed out that the model found by RARTS has 3.2M parameters, which is smaller than the 3.3M model found by DARTS. Moreover, RARTS outperforms other recent differentiable methods in accuracy and search cost at comparable model size. We also notice that the variance of RARTS performance is lower than that of DARTS. 
RARTS has also {\it arrested  architecture collapse} and only selected one skip-connection, as shown in Fig. \ref{arch}.
We are aware that different values of hyperparameters in the RARTS search stage may impact the latency of the models found by RARTS. Table \ref{hyperlat} has listed the latency of several models with different hyperparameters. Here we use the baseline setting of latency weight = $2\times10^{-3}$, batch size = 64, learning rate = $3\times10^{-4}$, weight decay = $1\times10^{-3}$. We change the value of one hyperparameter and keep the others the same during each experiment, so that we can see how sensitive the resulting latency is to a specific hyperparameter. First, the result shows that a small batch size of 16 can impact the model's latency, whereas a batch size of 32 or 64 can lead to similar latency. This is a positive phenomenon, since we prefer larger batch size as it requires less training time. Among the other hyperparameters, it is clear that the only factor that could cause a significant difference is the latency weight. A latency weight of $2\times10^{-2}$ is so large that its model has only 60\% latency compared with the baseline. The model's latency is not sensitive to the other hyperparameters, as the latency is around 22.0, and varies within 10\% only. This finding is beneficial, since we can fix the latency level via fixing the latency weight and find the model with the best accuracy among the models of similar latency level via tuning the other hyperparameters.

\textbf{Comparisons on ImageNet.}
ImageNet \cite{deng2009imagenet,russakovsky2015imagenet} is composed of over 1.2 million training images and 5,000 test images from 1,000 object classes. The architecture which is built of the cells learned on CIFAR-10 is transferred to be learned on ImageNet-1000, producing the results in Table \ref{Ima}. Even if our experiments are performed on a GTX 1080 Ti whose maximum memory allows only a batch size of 128, our 25.9\% error rate outperforms those of DARTS and SNAS (batch size 128), and is also comparable to those of GDAS (batch size 128) and MiLeNAS. MiLeNAS among some other algorithms in Table 2 have been implemented on Tesla V100 with batch size 1024, a much higher end hardware than that in our experiments. This partly explains its lower accuracy occurrence (2.80) on CIFAR-10 but higher accuracy after transfer to ImageNet. Typically ImageNet is trained better on larger GPU's because of the larger batch size. 
ProxylessNAS has obtained high accuracy on both CIFAR-10 and ImageNet, but their models are much larger than the other methods. It has avoided transfer learning as the training cost is reduced via path sampling. 
Inheriting the building blocks from DARTS and ProxylessNAS, FairDARTS has penalized the neutral (close to 0.5) architecture parameters, but its high accuracy also benefits from the relaxation on the search space. Their normal cells contain less than 8 operations since the operations with architecture parameters lower than a preset threshold are eliminated. This explains their smaller model size and comparable accuracy. P-DARTS has devised a progressive method to increase the depth of search. Their work shows that deeper cells have better capability of representation, which is also an improvement on the search space. PC-DARTS as a sampling method has achieved the least searching cost and can be trained directly on ImageNet. These methods are complementary to our work which is purely on the differentiable search algorithm without  modifying the search space of DARTS.

\textbf{Comparisons on NATS-Bench.} For NATS-Bench, one has to search a block of 6 nodes from the search space of 5 different operations, including zero, skip-connection, $3\times3$ average pooling, $1\times1$ convolution or $3\times3$ convolution \cite{dong2021nats}. Therefore, it includes 15,625 different candidate architectures and any DARTS style methods can be adapted easily to its search space. NATS-Bench has measured each architecture's performance under the same training settings, and hence  fair comparisons can be made between the discovered architectures since no further evaluation is needed on the local machines. In our experiments, we set batch size = 64,
initial weight learning rate = 0.025, momentum = 0.9, weight
decay = 0.0005, initial alpha learning rate = 0.0003, alpha
weight decay = 0.001, number of epochs = 100. Table \ref{NATS} presents the search results of DARTS vs. RARTS on NATS- Bench. RARTS has surpassed both DARTS-1 and DARTS-2 in accuracy by more than 20\% on CIFAR-10 and 6\% on CIFAR-100. Besides its success in accuracy, RARTS has totally escaped from the architecture collapse issue, i. e., the architectures found by RARTS from NATS-Bench contain no skip-connections. On the contrary, both architectures found by DARTS-1 and DARTS-2 contain 100\% and 38.9\% (average of 3 runs) skip-connections on CIFAR-10 and CIFAR-100, respectively. It is clear that too many skip-connections resulting in architecture collapse will impact the performance of the models greatly.

    \begin{table}[t]
        \begin{center}
          \caption{Test errors of DARTS vs. RARTS on NATS-Bench search space. The results of DARTS on NATS-Bench are from \cite{dong2021nats}. Ratio = the number of skip-connections over the number of total operations in the discovered architecture. }  \label{NATS}
    \begin{tabular}{l l c c}
    \toprule
     \specialrule{0em}{1pt}{1pt}
    Dataset & Method & Error (\%) & Ratio (\%)\\
    \specialrule{0em}{1pt}{1pt}
    \midrule
     \specialrule{0em}{1pt}{1pt}
     & DARTS-1  & 40.16 & 100 \\
    CIFAR-10 & DARTS-2 & 34.62 &  100 \\ 
     & RARTS & \textbf{11.48} & \textbf{0} \\
        \specialrule{0em}{1pt}{1pt}
      \midrule
        \midrule
     \specialrule{0em}{1pt}{1pt}
     & DARTS-1  & 38.74 & 38.9 \\
    CIFAR-100 & DARTS-2 & 39.51 &  38.9 \\ 
     & RARTS & \textbf{32.37} & \textbf{0} \\     
     \specialrule{0em}{1pt}{1pt}
    \bottomrule
    \end{tabular}
    \end{center}
\end{table}

\subsection{Search for Width}
To search the width of the architecture (number of channels in convolutional layers), we follow the settings of Network Slimming \cite{liu2017learning}, by introducing scoring parameters $\alpha$ to measure channel importance. Denote the original feature map by $F_{i,j}$ and define the new feature map 
$  \tilde{F}_{i,j} = \alpha_{i,j}\, F_{i,j}$, 
where $(i,j)$ are the layer and channel indices. 
Multiplying a channel of output feature map by $\alpha$ is equivalent to multiplying the convolutional kernels connecting to this output feature map by the same $\alpha$. We prune a channel if the corresponding $\alpha$ is 0 or very small. 
The $\alpha_{ij}$'s are learnable architecture parameters 
independent of channel weights, and hence is considered to have similar roles to the architecture parameters in the case of searching topological architecture.  

Although such treatment of scoring parameters is much like that in Network Slimming \cite{liu2017learning}, we point out that the single level formulation of RARTS and the training algorithm to learn those scoring parameters are novel. The first difference is that Network Slimming trains both weight and architecture parameters on the whole (training and validation) data, unlike DARTS or RARTS, without using either dataset splitting or network splitting. Another key difference between RARTS pruning and Network Slimming is in the search algorithm, i.e., Network Slimming trains the weights and the architecture jointly in one step, while RARTS trains them in a three-step iteration. Moreover, Network Slimming has used batch normalization weights as the scoring parameters. We point out that we could still define such a set of learnable architecture parameters $\alpha$, even if the batch normalization operation is not contained in the architecture.

We also compare RARTS with TAS \cite{dong2019network}, which is another width search method based on differentiable NAS, relying on both continuous relaxation via feature maps of various sizes and {\it model distillation}. The first difference is on how the channel scoring parameters are applied to the feature maps. For TAS, the channel parameters are treated as probabilities of candidate feature maps, smoothed by Gumbel-Softmax. Then a subset of feature maps is sampled to alleviate the high memory costs. RARTS is much simpler in its formulation, as it is a dot product of the channel parameters with the filter to be pruned. The second key difference is the use of a training technique called Knowledge Distillation (KD) \cite{hinton2015distilling} by TAS to improve accuracy. There are some other NAS based methods for width search, or channel pruning \cite{he2018amc,liu2019metapruning} mentioned in Section \ref{relateb}. Noting that our formulation of the problem and the criterion to evaluate results are different, we emphasize that out progress is in fusion of a new search algorithm and the width search task.

When using RARTS to search for width, we follow the hyperparameters and settings of Network Slimming as well. That is, learning rate = 0.1, weight decay = 0.0001, epochs = 160 \cite{liu2017learning}. In Table \ref{net_slim}, RARTS
outperforms the un-pruned baseline, Network Slimming (NS) and TAS \cite{dong2019network}
by over 10\% error reduction on CIFAR-10. While TAS does not offer an option to specify the pruning ratio of channels (PRC), the pruning ratio of FLOPs is around $30 \%$ for NS (40\% PRC), RARTS (40\% PRC) and TAS. So the comparison is fair. On CIFAR-100, RARTS still leads NS at the same PRC. The gap is smaller as the baseline network is less redundant.
Our experimental results reveal that the accuracy of TAS with KD is lower than (on CIFAR-10) or similar to (on CIFAR-100) that of RARTS, while TAS without the training technique like KD is 2\% worse \cite{dong2019network}. This supports the fact that RARTS works better as a differentiable method for width search, without regard to any other training tricks. Apart from the comparisons with the above methods, we also consider a pruning task for comparing DARTS and RARTS, which can be viewed as an ablation study of RARTS on the width search task. For this task, we prune MobileNetV2 \cite{sandler2018mobilenetv2} on a randomly sampled 20-class subset of ImageNet-1000, with $\ell_1$ regularization but unfixed pruning ratio. The pruning ratio can be learned automatically by the strong regularization term, as many of the architecture parameters are simply zero. Table \ref{rand} shows that RARTS also beats both random pruning and DARTS in accuracy. Even though the 2nd DARTS obtains a higher sparsity, it sacrifices the accuracy. 

\begin{table}
    \begin{center}
            \caption{Application of RARTS to ResNet-164 (baseline, 1.7 M parameters) channel pruning on CIFAR-10 and CIFAR-100, in comparison with the baseline, TAS and Network Slimming. 
            The numbers in the parentheses indicate the pruning ratio of channels (PRC). For NS and RARTS, PRC is fixed at 40\% or 60\%. 
            NS = Network Slimming. 
    }       \label{net_slim}
    \begin{tabular}{c l c}
    \toprule
    \specialrule{0em}{1pt}{1pt}
    Data & Method & Test Error (\%) 
    \\
    \specialrule{0em}{1pt}{1pt}
    \midrule
    \specialrule{0em}{1pt}{1pt}
    \multirow{5}{*}{\makecell{CIFAR-10}} & Baseline \cite{liu2017learning} & 5.42 
    \\
    & TAS \cite{dong2019network} & 6.00\\    
    & NS (40\% PRC) \cite{liu2017learning} & 5.08 \\
    & RARTS (40\% PRC) & \textbf{4.58}\\

    & NS (60\% PRC) \cite{liu2017learning} & 5.27 \\
    & RARTS (60\% PRC) & \textbf{4.90}
    
    \\

      \specialrule{0em}{1pt}{1pt}
      \midrule
      \specialrule{0em}{1pt}{1pt}
      \multirow{5}{*}{\makecell{CIFAR-100}} &  Baseline \cite{liu2017learning} & 23.37 
    \\
    & TAS \cite{dong2019network} & 22.24\\
    & NS (40\% PRC) \cite{liu2017learning} & 22.87 \\
    & RARTS (40\% PRC) & \textbf{22.64}\\
    & NS (60\% PRC) \cite{liu2017learning} & 23.91 \\
    & RARTS (60\% PRC) & \textbf{23.26}
    \\

    \specialrule{0em}{1pt}{1pt}
    \bottomrule
    \end{tabular}
    \end{center}
\end{table}

\begin{table}
    \begin{center}
            \caption{Application of RARTS to MobileNetV2 pruning on the ImageNet-R dataset (a randomly sampled subset of ImageNet-1000, with 20 object classes), compared with the baseline, random pruning, 1st and 2nd-order DARTS. Here random pruning means that we zero out channels randomly in accordance with the pruning ratio of RARTS. Average of 5 runs. PRC = the average pruning ratio of channels over the pruned layers. We note that the PRC can be high because the dataset is much smaller. 
    }       \label{rand}
    \begin{tabular}{l c c}
    \toprule
    \specialrule{0em}{1pt}{1pt}
    Method & Test Error. (\%) & PRC (\%)\\
    \specialrule{0em}{1pt}{1pt}
    \midrule
    \specialrule{0em}{1pt}{1pt}
    Baseline & 12.3 $\pm$ 1.4 & - \\
    Random Pruning & 12.0 $\pm$ 1.1 & 71.2 $\pm$ 1.9\\
    DARTS-1 & 10.1 $\pm$ 2.0 & 69.0 $\pm$ 0.9\\
    DARTS-2 & 9.8 $\pm$ 1.7 & 72.6 $\pm$ 2.0 \\
    RARTS & \textbf{8.2 $\pm$ 1.9} & 71.2 $\pm$ 1.9\\
    \specialrule{0em}{1pt}{1pt}
    \bottomrule
    \end{tabular}
    \end{center}

\end{table}

\section{Conclusion}

We have developed RARTS, a novel relaxed differentiable method for neural architecture search. We have proved its convergence theorem and compared it with DARTS on an analytically solvable model. Thanks to the design of data and network splitting, RARTS has achieved high accuracy and search efficiency over the state-of-the-art differentiable methods, especially DARTS, with a wide range of experiments, including both topology search and width search. These results support RARTS to be a more reliable and robust differentiable neural architecture search tool for various datasets and search spaces. In future work, we plan to incorporate search space sampling and regularization techniques to accelerate RARTS (as seen in several recent variants of DARTS) for broader applications in deep learning.

\section*{Acknowledgment}
The authors would like to thank the associate editor and the
anonymous referees for their careful reading and helpful
feedback, which improved the presentation of the paper.
The authors would also like to express their appreciation to
Dr. Shuai Zhang and Dr. Jiancheng Lyu for wonderful discussions
related to the project.


\bibliographystyle{IEEEtran}
\bibliography{main}

\begin{IEEEbiography}[{\includegraphics[width=1in,height=1.25in,clip,keepaspectratio]{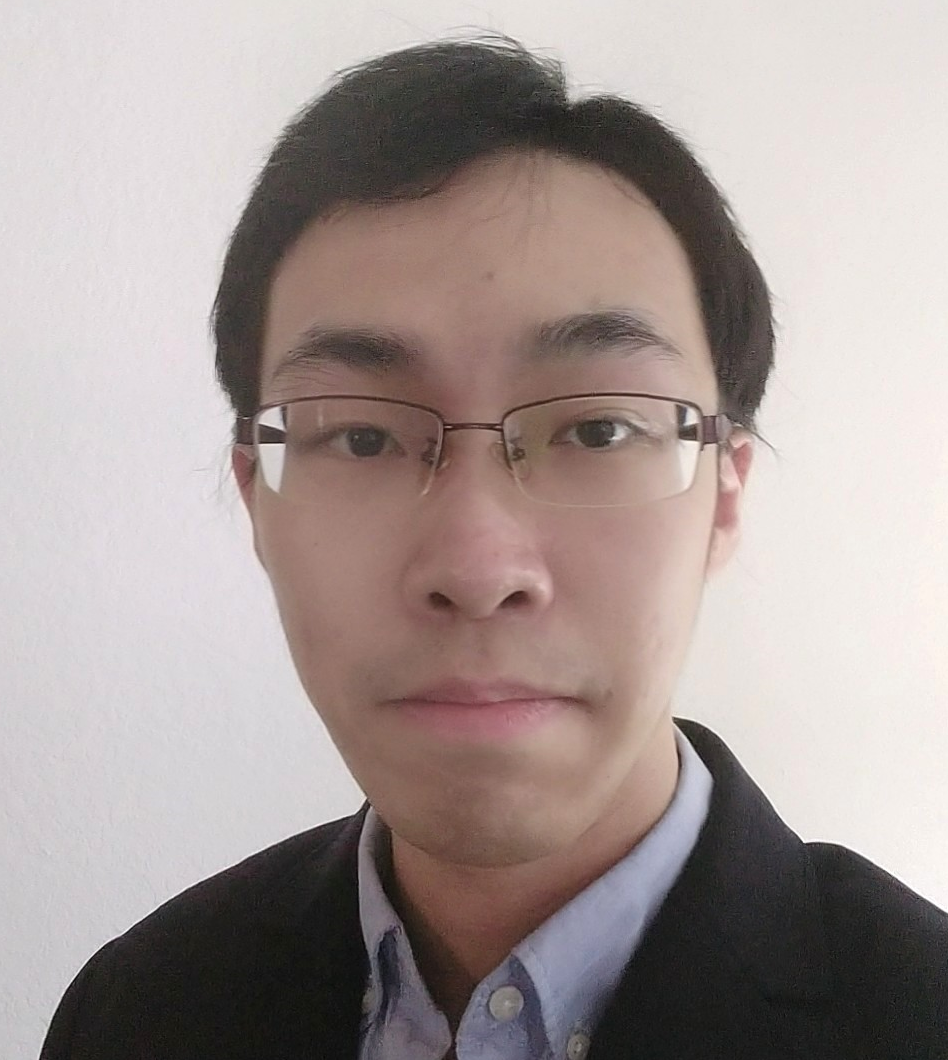}}]{Fanghui Xue} received the B.S. degree in mathematics from Fudan University, Shanghai, China, in 2016 and the M.S. degrees in analytics and mathematical risk management from Georgia State University, Atlanta, GA, in 2018. In May, 2022, he received his Ph.D. degree in mathematics at
University of California, Irvine, CA. His research interests include deep learning and computer vision, with a focus on AutoML and constructing efficient neural networks.

\end{IEEEbiography}

\begin{IEEEbiography}[{\includegraphics[width=1in,height=1.25in,clip,keepaspectratio]{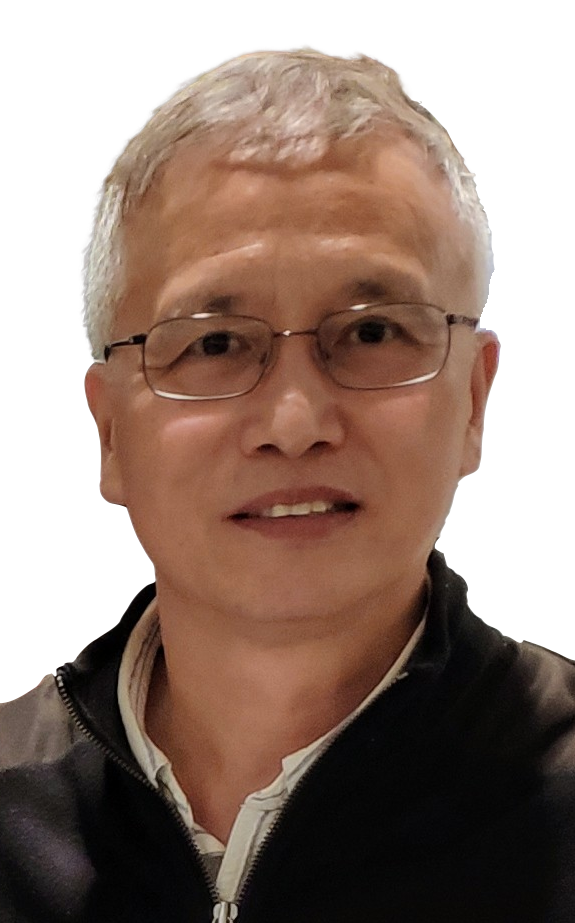}}]{Yingyong Qi} received the Ph.D. degree in
speech and hearing sciences from Ohio State Uni-
versity, in 1989, and the Ph.D. degree in electrical
and computer engineering from the University of
Arizona, in 1993. He held a faculty position at the
University of Arizona, from 1989 to 1999. He was
a Visiting Scientist at the Research Laboratory of
Electronics, Massachusetts Institute of Technol-
ogy, from 1995 to 1996, and a Visiting Scientist
at the Visual Computing Laboratory of Hewlett
Packard, Palo Alto, in 1998. He is currently a Senior Director of technology
at Qualcomm and a Researcher of the Department of Mathematics, Univer-
sity of California at Irvine. He has published over 100 scientific articles
and U.S. patents during his tenure at university and industry. His research
interests include speech processing, computer vision, and machine learning.
He received Klatt Memorial Award in Speech Science from the Acoustical
Society of America, in 1991, the First Award from the National Institute
of Health, in 1992, and the AASFAA Outstanding Faculty Award from the
University of Arizona, in 1998. More recently, he led a team winning the 3rd
Place of IEEE Low Power Image Recognition Competition, sponsored by
Google at CVPR 2019.
\end{IEEEbiography}

\begin{IEEEbiography}[{\includegraphics[width=1in,height=1.25in,clip,keepaspectratio]{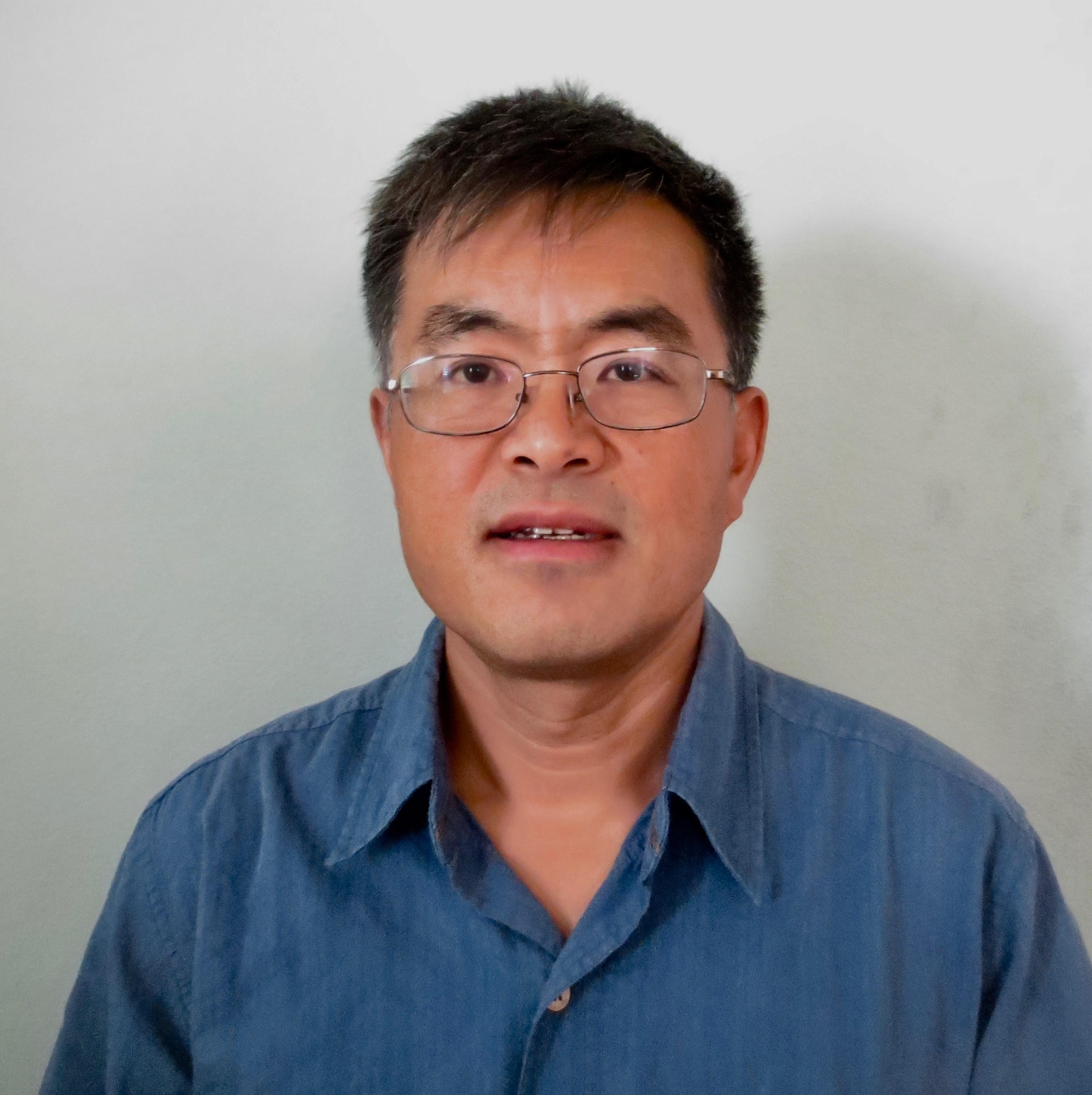}}]{Jack Xin} received the Ph.D. degree in mathematics from New York University’s Courant Institute of Mathematical Sciences, in 1990. He was a faculty at the University of Arizona, from 1991 to 1999, and the University of Texas at Austin, from 1999 to 2005. He is currently a Chancellor’s Professor of mathematics at UC Irvine. His research interests include applied analysis and computational methods, and their applications in multi-scale problems and data science. He is a fellow of Guggenheim Foundation, American Mathematical Society, American Association for the Advancement of Science, and the Society for Industrial and Applied Mathematics. He was a recipient of Qualcomm Faculty Award (2019–2022).

\end{IEEEbiography}

\EOD

\end{document}